# GAS-MIL: Group-Aggregative Selection Multi-Instance Learning for Ensemble of Foundation Models in Digital Pathology Image Analysis

Peiran Quan[1], Zifan Gu[1], Zhuo Zhao[1], Qin Zhou[1], Donghan M. Yang[1], Ruichen Rong[1], Yang Xie[1,2,3], Guanghua Xiao[1,2,3*]

[1]Quantitative Biomedical Research Center, Department of Health Data Science and Biostatistics, Peter O'Donnell Jr. School of Public Health, The University of Texas Southwestern Medical Center, Dallas, TX 75390, United States

[2]Department of Bioinformatics, UT Southwestern Medical Center, Dallas, Texas, 75390, USA

[3]Simmons Comprehensive Cancer Center, UT Southwestern Medical Center, Dallas, Texas, 75390, USA

Corresponding author:

Guanghua Xiao

Quantitative Biomedical Research Center

Department of Health Data Science and Biostatistics

Peter O'Donnell Jr. School of Public Health

The University of Texas Southwestern Medical Center

Phone: 214-648-4003

guanghua.xiao@utsouthwestern.edu

# Abstract

Foundation models (FMs) have transformed computational pathology by providing powerful, general-purpose feature extractors. However, adapting and benchmarking individual FMs for specific diagnostic tasks is often time-consuming and resource-intensive, especially given their scale and diversity. To address this challenge, we introduce Group-Aggregative Selection Multi-Instance Learning (GAS-MIL), a flexible ensemble framework that seamlessly integrates features from multiple FMs, preserving their complementary strengths without requiring manual feature selection or extensive task-specific fine-tuning. Across classification tasks in three cancer datasets—prostate (PANDA), ovarian (UBC-OCEAN), and breast (TCGA-BrCa)—GAS-MIL consistently achieves superior or on-par performance relative to individual FMs and established MIL methods, demonstrating its robustness and generalizability. By enabling efficient integration of heterogeneous FMs, GAS-MIL streamlines model deployment for pathology and provides a scalable foundation for future multimodal and precision oncology applications.

**Keywords**: Pathology image analysis, foundation models, multi-instance learning, feature extraction

# Introduction

With advances in imaging technologies and the increasing availability of whole-slide images (WSIs), pathologists now have access to a large volume of high-resolution data capturing cell morphology, tissue architecture, and the surrounding tumor microenvironment[1,2]. These detailed visual features are critical for accurate diagnosis, prognostication, and therapeutic decision-making[3-6]. At the same time, artificial intelligence (AI) has shown remarkable promise in accelerating and augmenting the traditionally labor-intensive tasks of digital pathology analysis and annotation. Early efforts focused on task-specific models, including cancer subtyping[7-9], metastasis detection[10,11], treatment response prediction[12,13], and survival estimation[14,15]. More recently, however, there has been a growing shift toward general-purpose, large-scale modeling approaches that can either operate across multiple cancer types[16,17] or simultaneously perform diverse diagnostic and prognostic tasks within a single framework[18-21].

These larger models, often termed foundation models (FMs), have emerged as powerful tools in digital pathology, characterized by their complexity, scale, and ability to generalize across heterogeneous data. Trained on massive datasets to learn rich and transferable representations, foundation models can be adapted to diverse downstream applications through fine-tuning or few-shot learning[22-27]. Recent advances—including Prov-GigaPath[28], Phikon[29,30], UNI[31], CONCH[32], and PLIP[33]—highlight their potential across key digital pathology tasks such as image classification, object detection, and semantic segmentation. For instance, Prov-GigaPath utilizes the LongNet[34] architecture to analyze WSI, capturing both local and global features, enabling accurate cancer subtype classification. Phikon, using the iBOT[35] framework for self-supervised learning, extracts critical features from large-scale pathology slides, demonstrating strong generalization capabilities in biomarker prediction. The UNI model, based on the DINOv2[36] framework, emphasizes stability and versatility in multi-task environments, excelling in various pathology image tasks, particularly whole-slide analysis. In addition, multimodal models like PLIP and CONCH integrate pathology images with paired reports, further improving performance on interpretive tasks.

However, despite their improved generalizability, benchmarking studies of foundation models continue to reveal substantial performance variability across diagnostic tasks and tissue types[37]. Such variability arises from differences in model architectures, pre-training strategies, and pathological conditions[38]. In practice, deploying these models still requires considerable time and resources for task- or dataset-specific fine-tuning. To address this challenge, there has been growing interest in automated strategies for integrating multiple foundation models. Early attempts—such as feature fusion or weighted ensemble approaches[39]—have shown only modest gains, often failing to capture the full representations of each model and overlooking subtle but clinically relevant features. Therefore, the development of methods that can effectively combine foundation models while preserving their unique strengths is highly desirable to reduce the reliance on extensive fine-tuning and to advance more broadly applicable AI solutions in pathology.

Multi-Instance Learning (MIL) has emerged as a promising paradigm for computational pathology, particularly when region-level annotations are scarce[40-42]. Under this framework, each WSI is represented as a "bag" of smaller image patches ("instances"), with only a slide-level label available. This allows models to learn from diagnostically informative regions while preserving their relationship to the broader tissue context. Several MIL approaches have demonstrated the feasibility of this approach in pathology: early works such as WELDON[43] and Chowder[44] applied min-max pooling to highlight salient regions, while more recent approaches—attention-based AB-MIL[45] and transformer-based TRANS-MIL[46]— have improved the ability to capture fine-grained local features and their global interactions. Collectively, these advances highlight why MIL is particularly well-suited for pathology, where tumor heterogeneity and microenvironmental variations mean that diagnostically critical information is often concentrated in distinct, spatially distributed regions across a slide.

In this study, we present Group-Aggregative Selection Multi-Instance Learning (GAS-MIL), a flexible and scalable framework for integrating multiple FMs by preserving the complementary features extracted by each (Figure 1). Built on the MIL paradigm, GAS-

MIL automatically identifies nuanced, context-dependent local features and links them to the broader tissue state. The main contributions of this work include:

(1) **Development of GAS-MIL**, a novel ensemble framework that performs feature fusion and selection through a flexible min–max and alignment strategy.

(2) **Comprehensive evaluation across three public cancer datasets**—prostate (PANDA), ovarian (UBC-OCEAN), and breast (TCGA-BrCa)—demonstrating robust and consistent performance across diverse pathological tasks.

(3) **Benchmarking** against both individual foundation models and state-of-the-art MIL methods (attention- and transformer-based), establishing the generalizability and advantage of GAS-MIL.

# Results

**GAS-MIL outperforms individual FMs on prostate, ovarian, and breast cancer**

GAS-MIL demonstrates strong performance in both multi-label and binary classification across three major cancer types—prostate, ovarian, and breast—compared with six individual FMs: Phikon, UNI, Prov-GigaPath, CONCH, PLIP, and LVM-Med[47]. Across all test datasets, GAS-MIL achieved performance that was either superior to or on par with the best-performing individual models (Figure 2 and Supplementary Tables 1–3).

For International Society of Urological Pathology (ISUP) prostate cancer grading, GAS-MIL reached a balanced accuracy of 0.788, exceeding the second-best model (Phikon) by 4.7%. It also achieved a Quadratic Weighted Kappa of 0.962, 1.3% higher than Prov-GigaPath. The superior performance is consistent across other metrics, with 3.83% higher accuracy, 3.97% higher weighted F1 score, and 4.72% higher balanced accuracy. The second-best performing FM alternated between Phikon and Prov-GagaPath. In ovarian cancer subtyping, GAS-MIL outperformed the second-best individual FM across all metrics (Accuracy + 5.56%, Balanced Accuracy + 6.45%, Quadratic Weighted Kappa + 11.31%, Weighted F1 + 5.55%), with the strongest comparator alternating between Phikon and UNI depending on the metric. For breast cancer classification (infiltrating ductal vs. lobular carcinoma), GAS-MIL trailed slightly behind the top-performing model, Prov-GigaPath, by 0.61% in accuracy and 0.85% in weighted F1. However, it surpassed Prov-GigaPath by 0.82% in AUC, highlighting competitive performance even when not the outright leader. Collectively, these results underscore GAS-MIL's robustness across diverse datasets and tasks.

We also compare these FMs with traditional Convolutional Neural Networks (CNN), including ResNet50 and two EfficientNet variants (Supp. Table 1). Interestingly, CNNs often outperformed several individual FMs but not GAS-MIL. For instance, in the PANDA dataset for prostate cancer, CONCH, LVM-Med, and PLIP consistently ranked below all three CNNs. Within CNNs, EfficientNet-B7 slightly outperformed EfficientNet-B1 and ResNet50. These findings suggest that while foundation models offer broad generalization, conventional architectures—when tuned to a specific task—can still be highly competitive. Nonetheless, GAS-MIL retained the overall advantage.

**Flexibility of GAS-MIL in feature alignment and FM composition**

GAS-MIL provides a flexible framework that supports varying numbers of foundation models. We observed that increasing the number of models in the ensemble generally improved balanced accuracy, quadratic weighted kappa, and weighted F1 score, while also reducing variance (Figure 3a–c). Importantly, this reduction in variance is not simply due to the number of available model combinations. For instance, the largest number of combinations occurs at three models (6 choose 3), yet variance continues to narrow as the ensemble size increases. This indicates that the observed stability reflects genuine benefits of integrating additional models rather than a combinatorial artifact. Notably, integrating only the top three individually best-performing models was sufficient to achieve near-optimal results. For instance, in terms of quadratic weighted kappa, the top three–model ensemble performed only 0.16% lower than a six-model ensemble (Figure 3d). These findings suggest that in practical applications, assembling a small number of strong models can deliver performance comparable to larger ensembles.

GAS-MIL also accommodates different feature extraction strategies, such as MLP or Attention-based alignment. In the UBC-OCEAN dataset, we tested 120 model combinations incorporating both approaches. Among the top five models, three used Attention; among the top ten, seven used Attention; and among the top 20, 13 used Attention (Supp. Figure 1). This preference for Attention was also evident across tasks: of the four overall top-performing GAS-MIL models, three employed Attention blocks for feature extraction, while only one used an MLP block (PANDA dataset). These results highlight Attention's advantage in capturing both global and local features, leading to stronger classification performance in many contexts.

**Ablation study demonstrates GAS-MIL's advantages over other MIL models**

To benchmark GAS-MIL against state-of-the-art MIL methods (AB-MIL, TRANS-MIL, and Chowder), we used features extracted by Phikon and UNI (methods). GAS-MIL achieved at least a 2% improvement with balanced accuracy (0.829) and weighted F1 score (0.826) compared to the second-best models, which are either Chowder or AB-MIL depending on the metric. TRANS-MIL performed the worst, suggesting that it may be less effective at modeling instance relevance within bags.

We further compared models trained on combined versus single FM features when classifying prostate cancer grade (Supp. Table 1). MIL models leveraging both Phikon and UNI consistently outperformed those using either FM alone. For example, AB-MIL with Phikon & UNI features improved over AB-MIL with UNI alone by 3.89% in accuracy, 3.27% in balanced accuracy, 0.81% in quadratic weighted kappa, and 3.56% in weighted F1. These results highlight the effectiveness of multi-model feature integration.

## Discussion

As foundation models become increasingly powerful and comprehensive, combining them to leverage their complementary strengths represents a promising strategy for advancing disease diagnostics. In this study, we developed GAS-MIL, an ensemble framework that integrates features generated by multiple foundation models for weakly supervised slide-level classification. GAS-MIL significantly improved key performance indicators—including accuracy, balanced accuracy, quadratic weighted kappa, and weighted F1 score—by employing feature fusion and selection mechanisms. These gains can be attributed to GAS-MIL's flexible design, particularly its min–max layer and feature alignment strategy, which enable the model to automatically detect informative local features while adaptively merging insights from diverse models.

The performance of individual foundation models further highlights the need for such an integrative framework. The DINO-based architectures (iBOT, DINOv2)—represented by Phikon, UNI, and Prov-GigaPath—show similar performance across the datasets, with less than 1% difference on the prostate ISUP grading task (PANDA). Notably, Phikon was trained on approximately 460 million pathology tiles, while Prov-GigaPath was trained on 1.3 billion slides, suggesting that simply increasing training set size does not always translate to improved performance. Since all three datasets were publicly available, these models may have already incorporated much of this information during pretraining; in contrast, a private dataset of rare disease cases could reveal performance differences linked to training size, though this question is beyond the scope of the current study.

Vision–language multimodal models (CONCH and PLIP) performed less strongly, with CONCH consistently surpassing PLIP. LVM-Med, trained on diverse medical image modalities, showed the weakest performance. For example, in the UBC-OCEAN dataset, its quadratic weighted kappa of 0.531 is 12% lower than that of CONCH, highlighting its limitations in handling pathology images. Taken together, these patterns underscore the importance of domain-specific pretraining for pathology and reinforce the value of an ensemble strategy like GAS-MIL, which can draw complementary strengths from otherwise heterogeneous models.

Leveraging this complementarity, we further observed that integrating multiple foundation models into GAS-MIL improves performance, although gains plateau beyond a certain ensemble size. For instance, in the PANDA dataset, a combination of three top-performing models (Phikon, UNI, and Prov-GigaPath) achieves results comparable to an ensemble of six models. This suggests that an optimized selection of a few models can deliver near-optimal performance without the added computational complexity of integrating more models. The ability to balance simplicity with performance gains offers practical advantages in clinical applications, where reducing computational overhead without sacrificing accuracy is essential. Moreover, GAS-MIL provides flexibility in selecting feature extraction mechanisms, such as MLP or Attention blocks. Our experiments (Supp. Figure 1) demonstrate that models using Attention blocks outperform those using MLP blocks, particularly in complex datasets like UBC-OCEAN, where attention-based models better capture global and local features necessary for accurate diagnostics.

In conclusion, GAS-MIL represents a significant advancement in computational pathology, offering a highly flexible and efficient framework for weakly supervised classification tasks. As the field shifts toward building increasingly large and complex FMs, selecting the most suitable model for a specific task remains challenging without extensive and time-consuming benchmarking. By integrating multiple foundation models and employing effective feature extraction mechanisms, GAS-MIL delivers robust and consistent results across diverse datasets, with internal fine-tuning achieved through MLP or Attention-based feature alignment. Importantly, GAS-MIL offers a scalable approach to perform task-specific tasks without finetuning FMs, paving the way for more precise and generalizable computational pathology workflows.

## Materials and Methods

**Dataset**

**Prostate Cancer Dataset (PANDA)[48].** The PANDA (The Prostate cANcer graDe Assessment) dataset includes 10,616 digitized H&E-stained WSIs of prostate biopsies, making it the largest publicly available dataset on prostate cancer. We use this dataset to predict the ISUP (International Society of Urological Pathology) score on prostate tissues[49]. This score, derived from the Gleason grading system, is divided into five levels based on severity, with 0 representing normal tissue. It plays a crucial role in assessing tumor aggressiveness and guiding treatment decisions for prostate cancer patients. These images are sourced from two institutions: 51% of slides from Karolinska Institute and 49% from Radboud University Medical Center. Each slide in the dataset has a full-resolution magnification of 20X, with a pixel size of approximately 0.5 micrometers. Additional image levels are available at 5X and 1.25X magnifications. Generally, each slide contains one biopsy needle of tissue; however, some images from Karolinska include two sections of the same biopsy needle. This dataset provides a rich resource for developing and testing computational pathology tools. We follow the approach used by the first-place solution of the Kaggle PANDA Challenge (https://github.com/kentaroy47/Kaggle-PANDA-1st-place-solution), which involved removing duplicates and noisy data, resulting in a clean set of 9,128 samples. We then randomly divide the data into 5392 for training, 1933 for validation, and 1803 for testing, using stratified sampling to maintain the class distribution across all sets.

**Ovarian Cancer Dataset (UBC-OCEAN)[50,51].** The UBC-OCEAN (Ovarian Cancer subtypE clAssification and outlier detectioN) is the world's most extensive ovarian cancer dataset of histopathology images obtained from more than 20 medical centers. This dataset comprises high-resolution images of ovarian cancer subtypes, including WSIs (20X) and tissue microarrays (TMAs, 40X), aimed at developing AI models that can classify different ovarian cancer subtypes and detect outliers. The dataset includes 5 ovarian cancer subtypes like high-grade serous carcinoma (HGSC), clear-cell ovarian carcinoma (CC), endometrioid carcinoma (EC), low-grade serous carcinoma (LGSC), and mucinous carcinoma (MC). This classification is crucial for personalized medicine and targeted therapy, as each subtype exhibits distinct responses to treatment and biological characteristics. By utilizing the UBC-OCEAN dataset, we aim to improve the accuracy of ovarian cancer subtype classification and gain deeper insights into the tissue characteristics of each subtype, thereby supporting the diagnosis and treatment of ovarian cancer. The available dataset includes 536 samples, we randomly divide the data into 321 for training, 107 for validation, and 108 for testing.

**Breast Cancer Dataset (TCGA-BRCA)[52].** The TCGA-BRCA dataset is a subset of The Cancer Genome Atlas (TCGA) project, focusing on Breast Invasive Carcinoma (BRCA). The TCGA-BRCA dataset provides a wealth of histopathological slides for breast cancer research, covering samples from approximately 1,100 patients. In this study, our task is to classify Infiltrating Ductal Carcinoma (IDC) and Infiltrating Lobular Carcinoma (ILC). IDC and ILC exhibit significant differences in treatment response and metastasis patterns, and correctly classifying IDC and ILC is essential for accurate pathology diagnosis and personalized therapy. Our goal is to enhance the foundational models' performance in classifying IDC and ILC. We obtain 721 Infiltrating Ductal Carcinoma and 172 Infiltrating Lobular Carcinoma from 893 WSIs. We randomly select 580 training samples, 150 validation samples, and 163 testing samples from the TCGA-BRCA dataset. The dataset is divided based on the characteristics of the patients' tissue slides, ensuring a balanced distribution of sample categories and feature characteristics across the training, validation, and testing sets.

**GAS-MIL Model**

We aim to enhance model stability and robustness by integrating features generated from multiple foundation models without significantly increasing computational demands. We enhance Chowder's capabilities by increasing its network width and maintaining the Max-Min layer to optimize efficiency. Figure 1 in the provided description outlines our GAS-MIL algorithm workflow.

For a WSI, we randomly extract $n$ patches and use $K$ foundation models to extract features from each patch. The $k$-th foundation model ($k$ = 1, 2, 3…, $K$) produces feature vectors of dimension $m_k$. Consequently, the total feature dimension for all $K$ models on a single patch is given by $m = \sum_{k=1}^{K} m_k$. Thus, the feature map for the $k$-th foundation model across all patches, denoted as $\mathbf{A}_k$, has a size of $n \times m_k$. The overall feature map of the WSI, denoted as $\mathbf{A}$, is obtained by concatenating the feature maps of all $n$ patches, resulting in a size of $n \times m$. Further details are provided in Eq. (1).

Instead of traditional fusion operations, we propose a Grouped Feature Extraction Blocks (GFEB, Supplementary Note 1) architecture to fully leverage the potential of multi-model features. To effectively integrate different models, we first align $\boldsymbol{A}_k$ from $n \times m_k$ to $n \times c$ using MLPs or attention mechanisms, yielding $\boldsymbol{B}_k$, where $c$ represents the number of classes in the downstream tasks. Additionally, we align $\boldsymbol{A}$ from $n \times m$ to $n \times c$, obtaining $\boldsymbol{B}_{K+1}$, to better capture interactions across different models. This architecture accounts for both the individual contributions of foundation models and their combined influence, thereby enhancing feature extraction and integration. Further details are given in Eq. (2).

To focus on the most representative patches and regions while reducing computational costs, we utilize a Max-Min layer to process the feature maps from the previous step. This layer selects only the top $s$ largest and smallest values from each row of $B_k$, transforming their size from $n \times c$ to $2s \times c$ and forming features maps $C_k$. The Max-Min layer aims to capture both high-importance instances and regions strongly indicative of the class's absence, incorporating both positive and negative evidence. By filtering out the most relevant features, this approach prioritizes elements with the greatest impact on predictions while discarding those with lower significance. Additionally, during training, backpropagation is applied only to these selected tiles, significantly improving computational efficiency by reducing overhead. Further details are provided in Eq. (3).

Finally, we concatenate all $C_k$, then transpose the concatenated feature maps to obtain the final feature map $D$ for downstream classification. A classification head, consisting of linear, sigmoid, and dropout layers, is then applied to $D$ to perform the classification task. Further details are given in Eq. (4).

$$A = [A_1; \ldots; A_k; \ldots; A_K] \text{ where } A_k \in \mathbb{R}^{n \times m_k},\ k \in [1, \ldots, K] \tag{1}$$

$$B_k = GFEB_k(A_k) \text{ where } B_k \in \mathbb{R}^{n \times c}, k \in [1, \ldots, K+1] \tag{2}$$

$$C_k = MaxMin(B_k) \text{ where } C_k \in \mathbb{R}^{2s \times c}, k \in [1, \ldots, K+1] \tag{3}$$

$$D = [C_1; \ldots; C_k; \ldots; C_{K+1}]^T \text{ where } D \in \mathbb{R}^{c \times 2(K+1)s} \tag{4}$$

**Pre-Processing**

**Tissue Detection and Tiling**. In digital pathology, the size of WSI often poses a challenge as these images are typically very large and contain regions with minimal biological relevance. Our objective is to identify tissue areas for the extraction of smaller, more relevant samples for detailed analysis. To achieve this, we employ the Otsu thresholding method [53]. First, the WSI is loaded at a 1.25x magnification in the HSV color space, and Otsu thresholds are calculated for the Hue (H) and Saturation (S) channels. A mask is then created using a logical AND operation, which retains only the areas where both H and S values exceed their respective thresholds, set at 0.6 for this procedure. Finally, the mask is dilated and resized to 20x magnification level, and patches with size 224x224 pixels are cropped from WSI accordingly.

**Feature extraction**. After achieving the patches from WSIs, we feed them to the foundation models. For each WSI, a N×M feature map is generated, where n is the number of patches for this WSI, and m is the number of features per patch combined from all foundation models. For example, for Phikon and UNI ensemble, Phikon produces 768 features per patch, and UNI produces 1024, resulting in a combined total of 1792 features per patch. To reduce the feature dimensions and maintain computational efficiency, we randomly select 200 patches per WSI. For WSIs with fewer than 200 patches, we pad the feature set with zeros to reach 200 patches. The extracted features are then as the input for the for the GAS-MIL model. For detailed procedures about feature extraction, further information is available on the corresponding GitHub pages Phikon (https://github.com/owkin/HistoSSLscaling), UNI (https://github.com/mahmoodlab/UNI), Prov-GigaPath(https://github.com/prov-gigapath/prov-gigapath ), CONCH(https://github.com/mahmoodlab/CONCH ), PLIP(https://github.com/PathologyFoundation/plip ) and LVM-Med (https://github.com/duyhominhnguyen/LVM-Med ).

**Training**

For the GAS-MIL model, we use GFEB with grouped MLPs, where the hidden layer dimension is 192, or with grouped attention mechanisms, featuring a dimensionality of 512 for the features and 256 for the attention. In the Max-Min layer, we select S equals 20 for each of these K+1 groups, capturing both the top 20 and bottom 20 features. In the classification head, we set the hidden layer dimension to 96 and dropout probability to 0.3.

Our experiments are conducted on a Tesla V100 GPU with 32 GB of memory. The model is trained for 100 epochs with a batch size of 128, using the AdamW optimizer to optimize network weights. We set the initial learning rate at 0.001 with a weight decay parameter of 0.05. Additionally, to prevent overfitting, we employ Gaussian noise with a standard deviation of 1.5 to the inputs to simulate uncertainty or interference in the data during training. We also use an early stop mechanism, which stops training if the model performance does not improve on a validation set over a certain number of epochs. To ensure balanced training across classes each epoch, we utilize a weighted random sampling strategy. For ISUP grading task (PANDA), we consider ISUP grades as a multi-label task rather than a multi-class classification task and use Binary Cross-Entropy Loss (BCELoss) as the loss function. Because ISUP grades exhibit an ordinal relationship where higher grades correspond to greater cancer severity, multi-label methods effectively capture this sequential nature. Compared to multi-class classification, the multi-label approach can reduce the problem of the model incorrectly classifying a sample as a significantly different grade than its actual grade. For the other tasks, we use Cross-Entropy Loss as the loss function.

**Evaluation**

We use accuracy, balanced accuracy, quadratic weighted kappa and weighted F1 score to evaluate multiclass tasks and Accuracy, AUC, F1 score for binary classification.
**Accuracy** is one of the most straightforward evaluation metrics. It measures the ratio of correctly predicted instances to the total number of instances in the dataset.
**Balanced Accuracy** addresses the limitations of accuracy in the presence of class imbalance by considering the sensitivity of the classifier to each class. It calculates the arithmetic mean of sensitivity (true positive rate) across all classes.
**Quadratic Weighted Kappa** is a statistical measure of inter-rater agreement that assesses the agreement between two ratings, considering the possibility of chance agreement. It is commonly used to evaluate the agreement between predicted and actual class labels. Unlike simple accuracy, quadratic weighted kappa considers partial credit for partially correct predictions and penalizes disagreements more severely. It ranges from -1 (complete disagreement) to 1 (complete agreement), with 0 indicating agreement equivalent to chance. Higher values indicate better agreement between raters.
**Weighted F1 Score** is a weighted average of the F1 scores of each class, where the weight is proportional to the number of true instances for each class. It considers both precision and recall, providing a balance between them. The F1 score is the harmonic mean of precision and recall, and the weighted F1 score extends this concept to multi-class classification problems by considering the class distribution.

**AUC** (Area Under the Curve) is a performance metric commonly used for binary classification problems. It represents the area under the Receiver Operating Characteristic (ROC) curve, which plots the true positive rate (sensitivity) against the false positive rate (1-specificity) at various threshold settings. AUC measures the classifier's ability to distinguish between positive and negative classes, regardless of the class distribution. An AUC of 1 indicates perfect discrimination, while an AUC of 0.5 suggests random guessing. Higher AUC values indicate better overall model performance, particularly in handling class imbalances by assessing how well the model ranks positive instances higher than negative ones.

For model comparison, we utilize CONCH, Phikon, UNI, Prov-GigaPath, PLIP and LVM-Med for feature extraction. These extracted features are then utilized to train the chosen MIL model AB-MIL, following the methodologies outlined in the respective publications of the foundational models and MIL model. For additional comparison, on ISUP grading task (PANDA), we also compare with classic CNN models (ResNet-50[54], EfficientNet B1[55], EfficientNet B7[55]). We use ResNet-50 as the base model. EfficientNet B1 which has demonstrated effectiveness in the Kaggle PANDA challenge, while EfficientNet B7 boasts a larger model size and depth compared to EfficientNet B1.

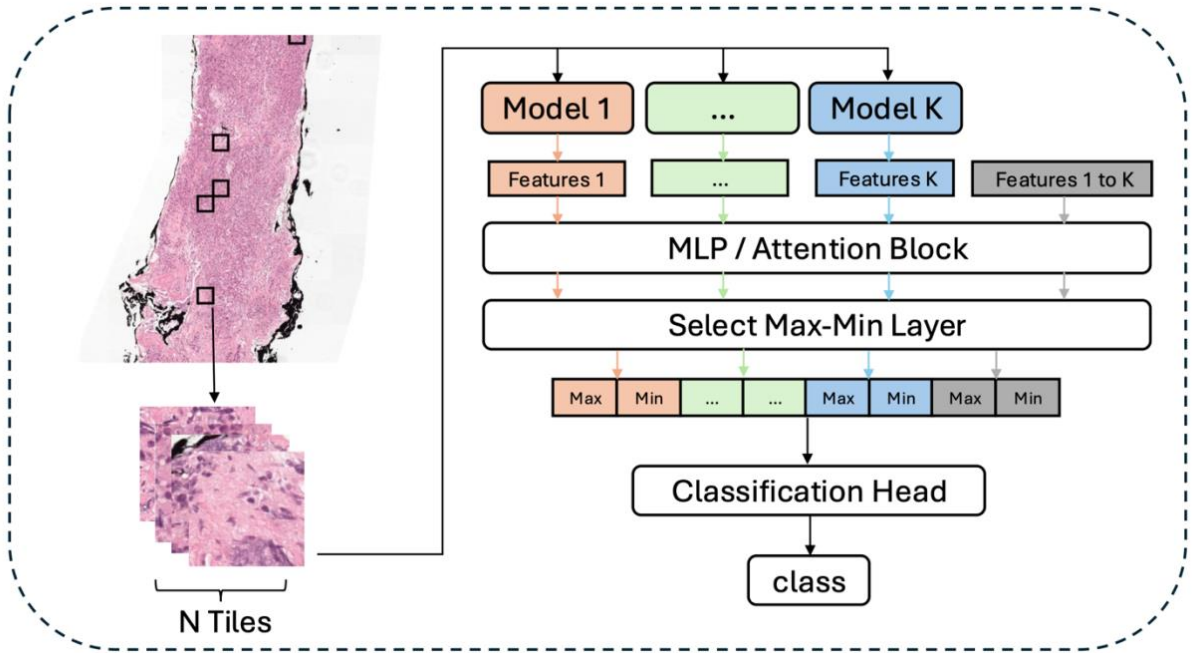

**Figure 1**. GAS-MIL pipeline. Tissue areas are first detected from the WSI, and patches are extracted accordingly. Next, features are generated using *k* foundation models and combined. These combined features are then processed through a Grouped MLP or an attention block to select maximal and minimal features. Finally, the selected features are merged and classified using a classification head.

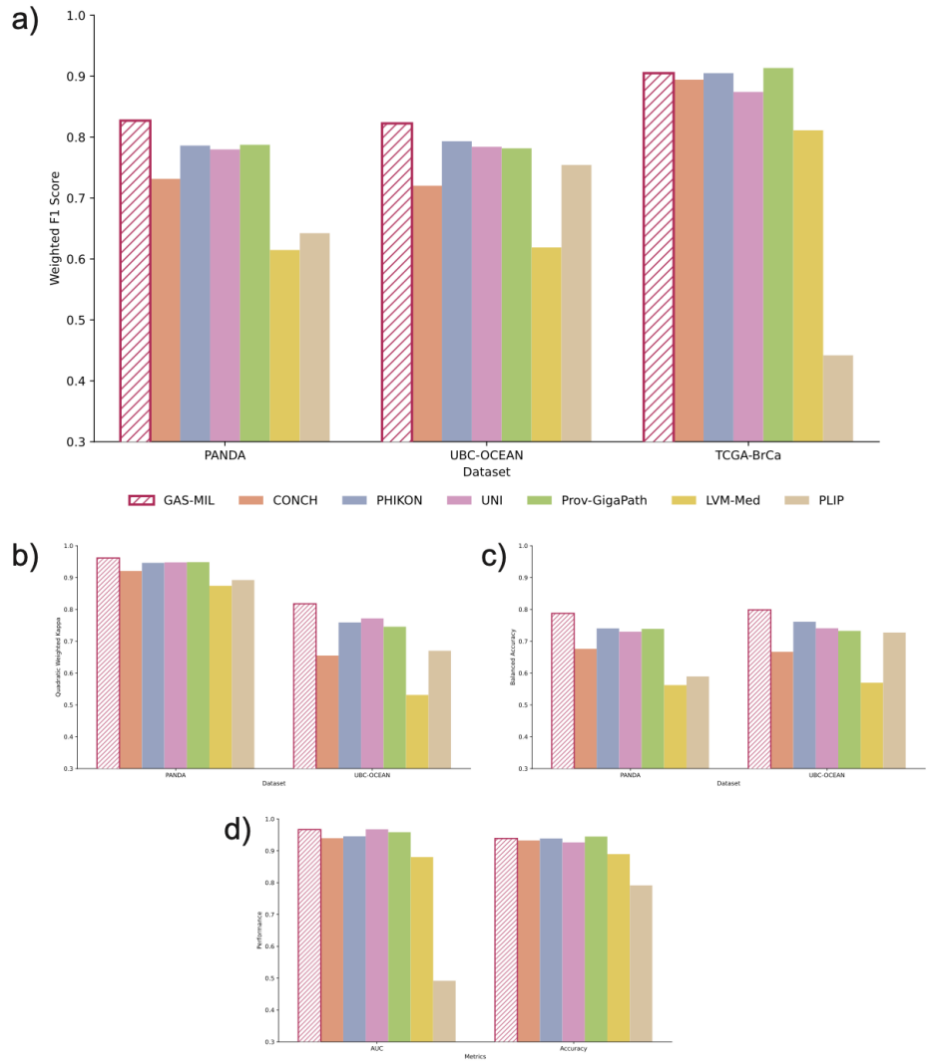

**Figure 2. GAS-MIL performance across multiple cancer types.** (a) Comparison of GAS-MIL and its comparisons across prostate (PANDA), ovarian (UBC-OCEAN), and breast cancer (TCGA-BrCa) slide-level classification tasks. (b-c) Performance of GAS-MIL in the PANDA and UBC-OCEAN testing set, evaluated by b) Quadratic Weighted Kappa and c) Balanced Accuracy. (d) Performance on the TCGA_BRCA dataset, evaluated by AUC and Accuracy.

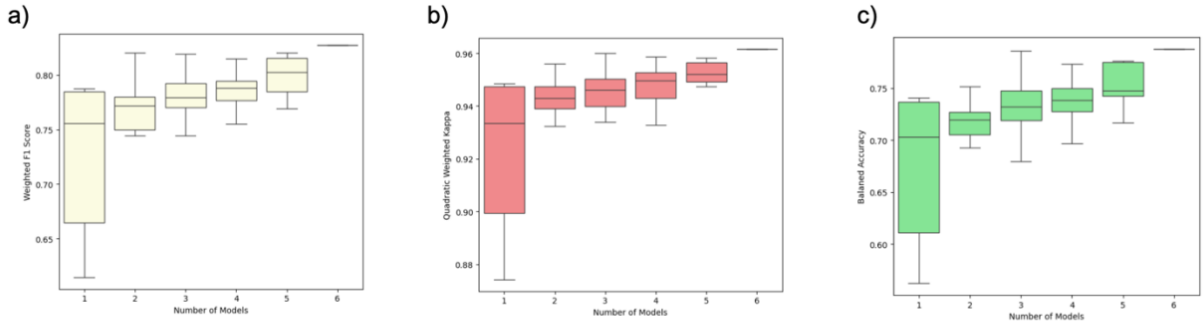

| Models Compositions | Accuracy | Balanced Accuracy | Quadratic Weighted Kappa | Weighted F1 Score |
|---|---|---|---|---|
| Phikon & UNI | 0.8286 | 0.7899 | 0.9605 | 0.8264 |
| Prov-GigaPath & Phikon & UNI | **0.8314** | **0.7909** | 0.9599 | **0.8304** |
| LVM-Med & PLIP & Phikon & UNI | 0.8159 | 0.7868 | 0.9581 | 0.8151 |
| LVM-Med & Prov-GigaPath & Phikon & UNI & CONCH | 0.8236 | 0.7763 | 0.9582 | 0.8202 |
| LVM-Med & Prov-GigaPath & Phikon & UNI & CONCH & PLIP | 0.8275 | 0.7876 | **0.9615** | 0.8271 |

**Figure 3. Impact of foundation model composition on GAS-MIL performance.** (a–c) Performance of GAS-MIL on the PANDA test set with varying numbers of foundation models, measured by (a) weighted F1 score, (b) quadratic weighted kappa, and (c) balanced accuracy. (d) Best-performing ensemble configurations for each combination size (n = 2–6).

Table 1. Comparison of four MIL models on the PANDA test set. All MIL models use features extracted from Phikon and UNI. MIL: Multiple instance learning.

| MIL | Accuracy | Balanced Accuracy | Quadratic Weighted Kappa | Weighted F1 Score |
| --- | --- | --- | --- | --- |
| AB-MIL | 0.8131 | 0.7540 | 0.9589 | 0.8073 |
| TRANS-MIL | 0.7787 | 0.7080 | 0.9525 | 0.7764 |
| Chowder | 0.8170 | 0.7628 | 0.9559 | 0.8154 |
| GAS-MIL | **0.8286** | **0.7899** | **0.9605** | **0.8264** |

**Supplementary Note 1. The Grouped Feature Extraction Blocks (GFEB) architecture.**

GFEB consists of $K+1$ feature extraction blocks (MLPs or Attentions). For MLP-based blocks, each consists of two linear layers with Sigmoid activation functions. For Attention-based blocks, the input features are first projected to the target feature dimension through a linear layer. Then, a scaled dot-product attention mechanism is applied, which computes positional relevance scores between the input features using linear projections of Queries, Keys, and Values. The attention weights are normalized using the SoftMax function and subsequently used to compute a weighted sum of the value vectors, producing the attention output. Finally, the attention output is obtained through a linear layer.

Supplementary Table 1. Comparison of 6 foundation models, 3 CNNs, and and GAS-MIL on the PANDA testing set.

| Models | Accuracy | Balanced Accuracy | Quadratic Weighted Kappa | Weighted F1 Score |
|---|---|---|---|---|
| ResNet-50 | 0.7249 | 0.6879 | 0.9357 | 0.7270 |
| EfficientNet B1 | 0.7321 | 0.6935 | 0.9324 | 0.7350 |
| EfficientNet B7 | 0.7499 | 0.6957 | 0.9390 | 0.7502 |
| CONCH | 0.7338 | 0. 6762 | 0. 9207 | 0. 7315 |
| Phikon | 0.7865 | 0.7404 | 0. 9463 | 0. 7861 |
| UNI | 0.7781 | 0. 7301 | 0. 9478 | 0. 7798 |
| Prov-GigaPath | 0.7892 | 0.7389 | 0.9485 | 0.7874 |
| LVM-Med | 0.6151 | 0.5622 | 0.8742 | 0.6147 |
| PLIP | 0.6417 | 0.5894 | 0.8923 | 0.6423 |
| GAS-MIL* | **0.8275** | **0.7876** | **0.9615** | **0.8271** |

Supplementary Table 2. Comparison of 6 foundation models and GAS-MIL on the UBC-OCEAN testing set.

| Models | Accuracy | Balanced Accuracy | Quadratic Weighted Kappa | Weighted F1 Score |
|---|---|---|---|---|
| CONCH | 0.7315 | 0.6668 | 0.6550 | 0.7202 |
| Phikon | 0.7963 | 0.7613 | 0.7591 | 0.7932 |
| UNI | 0.7870 | 0.7408 | 0.7718 | 0.7841 |
| Prov-GigaPath | 0.7870 | 0.7326 | 0.7457 | 0.7816 |
| LVM-Med | 0.6481 | 0.5697 | 0.5313 | 0.6189 |
| PLIP | 0.7593 | 0.7273 | 0.6701 | 0.7543 |
| GAS-MIL | 0.8241 | 0.7986 | 0.8177 | 0.8225 |

Supplementary Table 3. Comparison of 6 foundation models and GAS-MIL on the TCGA-BrCa testing set.

| Models | Accuracy | AUC | F1 Score |
|---|---|---|---|
| CONCH | 0.9325 | 0.9396 | 0.8943 |
| Phikon | 0.9387 | 0.9455 | 0.9050 |
| UNI | 0.9264 | 0.9676 | 0.8742 |
| Prov-GigaPath | 0.9448 | 0.9585 | 0.9135 |
| LVM-Med | 0.8896 | 0.8803 | 0.8112 |
| PLIP | 0.7914 | 0.4918 | 0.4418 |
| GAS-MIL | 0.9387 | 0.9667 | 0.9050 |

Supplementary Figure 1. Comparison of GAS-MIL models using MLP versus Attention mechanisms on the UBC-OCEAN dataset, showing the distribution of top-n performing models (n = 5, 10, 20).

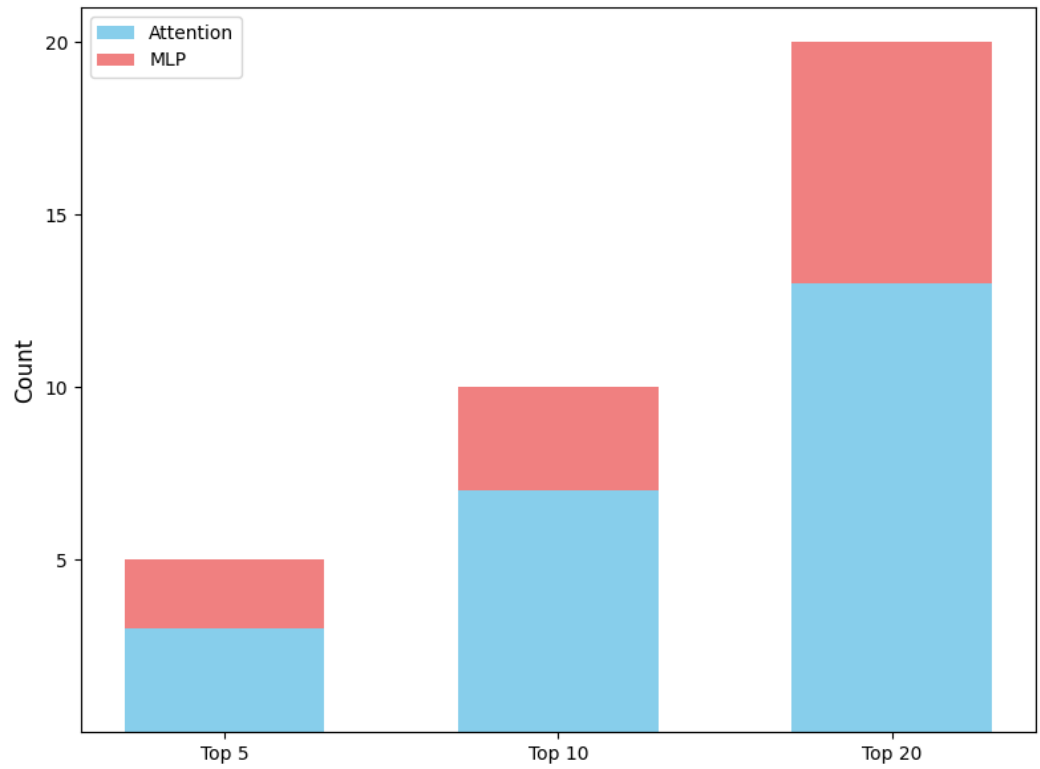